\documentclass[11pt,letterpaper]{article}
\usepackage{emnlp2016}
\usepackage{tablefootnote}
\usepackage{times}
\usepackage{latexsym}
\usepackage{dirtytalk}
\usepackage{pgfplots}
\usepackage{placeins}
\usepackage{amsmath}
\usepackage{placeins}
\usepackage{subcaption}
\usepackage[]{algorithm2e}
\graphicspath{ {Figures/} }
\emnlpfinalcopy 
\def\emnlppaperid{1100} 

\title{Character-Level Question Answering with Attention}

\author{
David Golub \\
University of Washington\\
\tt{golubd@cs.washington.edu} \\
\And
Xiaodong He \\
Microsoft Research \\
\tt{xiaohe@microsoft.com} \\
}

\date{}

\begin{document}

\maketitle

\begin{abstract}
We show that a character-level encoder-decoder framework can be successfully applied to question answering with a structured knowledge base. We use our model for single-relation question answering and demonstrate the effectiveness of our approach on the SimpleQuestions dataset \cite{babi_dataset}, where we improve state-of-the-art accuracy from 63.9\% to 70.9\%, without use of ensembles. Importantly, our character-level model has 16x fewer parameters than an equivalent word-level model, can be learned with significantly less data compared to previous work, which relies on data augmentation, and is robust to new entities in testing. \end{abstract}

\section{Introduction}

\begin{figure*}[model_architecture]
\hspace*{0cm}\includegraphics[width=17cm]{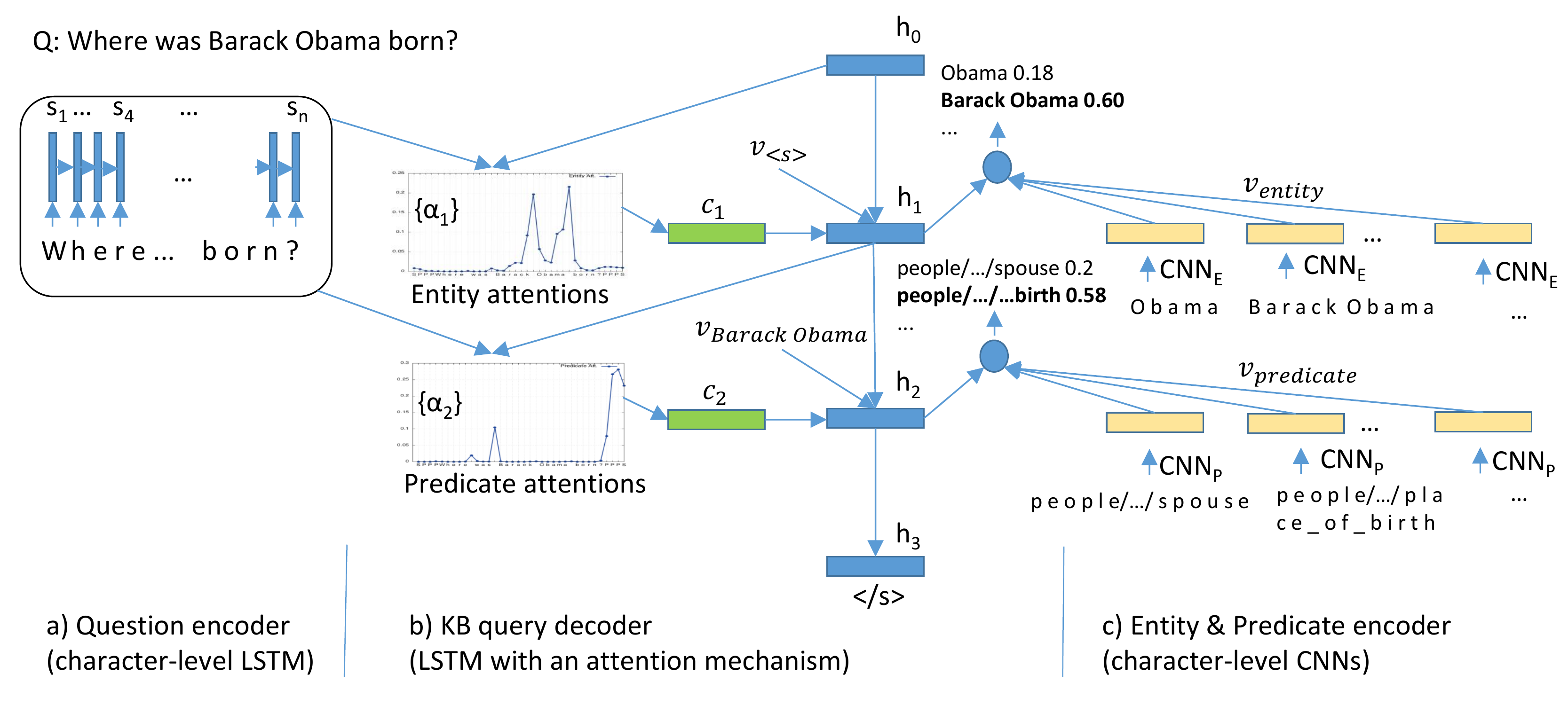}
\caption{Our encoder-decoder architecture that generates a query against a structured knowledge base. We encode our question via a long short-term memory (LSTM) network and an attention mechanism to produce our context vector. During decoding, at each time step, we feed the current context vector and an embedding of the English alias of the previously generated knowledge base entry into an attention-based decoding LSTM to generate the new candidate entity or predicate.}
\end{figure*}

Single-relation factoid questions are the most common form of questions found in search query logs and community question answering websites \cite{qa_acl,fader2013paraphrase}. A knowledge-base (KB) such as Freebase, DBpedia, or Wikidata can help answer such questions after users reformulate them as queries. For instance, the question \say{Where was Barack Obama born?} can be answered by issuing the following KB query:
\begin{align*}
\lambda(x).place\_of\_birth(Barack\_Obama, x)
\end{align*} 
However, automatically mapping a natural language question such as \say{Where was Barack Obama born?} to its corresponding KB query remains a challenging task. 

There are three key issues that make learning this mapping non-trivial. First, there are many paraphrases of the same question. Second, many of the KB entries are unseen during training time; however, we still need to correctly predict them at test time. Third, a KB such as Freebase typically contains millions of entities and thousands of predicates, making it difficult for a system to predict these entities at scale \cite{qa_acl,Fader2014,babi_dataset}. In this paper, we address all three of these issues with a character-level encoder-decoder framework that significantly improves performance over state-of-the-art word-level neural models, while also providing a much more compact model that can be learned from less data.

First, we use a long short-term memory (LSTM) \cite{LSTM} encoder to embed the question. Second, to make our model robust to unseen KB entries, we extract embeddings for questions, predicates and entities purely from their character-level representations. Character-level modeling has been previously shown to generalize well to new words not seen during training \cite{tweet_char,chung2016character}, which makes it ideal for this task. Third, to scale our model to handle the millions of entities and thousands of predicates in the KB, instead of using a large output layer in the decoder to directly predict the entity and predicate, we use a general interaction function between the question embeddings and KB embeddings that measures their semantic relevance to determine the output. The combined use of character-level modeling and a semantic relevance function allows us to successfully produce likelihood scores for the KB entries that are not present in our vocabulary, a challenging task for standard encoder-decoder frameworks.

Our novel, character-level encoder-decoder model is compact, requires significantly less data to train than previous work, and is able to generalize well to unseen entities in test time. In particular, without use of ensembles, we achieve 70.9\% accuracy in the Freebase2M setting and 70.3\% accuracy in the Freebase5M setting on the SimpleQuestions dataset, outperforming the previous state-of-arts of 62.7\% and 63.9\% \cite{babi_dataset} by 8.2\% and 6.4\% respectively. Moreover, we only use the training questions provided in SimpleQuestions to train our model, which cover about 24\% of words in entity aliases on the test set. This demonstrates the robustness of the character-level model to unseen entities. In contrast, data augmentation is usually necessary to provide more coverage for unseen entities and predicates, as done in previous work \cite{babi_dataset,qa_acl}.

\raggedbottom
\section{Related Work}
Our  work is motivated by three  major  threads  of  research  in  machine learning and natural language processing: semantic-parsing for open-domain question answering, character-level language modeling, and encoder-decoder methods. 

Semantic parsing for open-domain question answering, which translates a question into a structured KB query, is a key component in question answering with a KB. While early approaches relied on building high-quality lexicons for domain-specific databases such as GeoQuery \cite{tang2001using}, recent work has focused on building semantic parsing frameworks for general knowledge bases such as Freebase \cite{qa_acl,subgraph_embeddings,babi_dataset,percy_liang,fader2013paraphrase}. 

Semantic parsing frameworks for large-scale knowledge bases have to be able to successfully generate queries for the millions of entities and thousands of predicates in the KB, many of which are unseen during training. To address this issue, recent work relies on producing embeddings for predicates and entities in a KB based on their textual descriptions \cite{subgraph_embeddings,babi_dataset,qa_acl,Yih2015SemanticPV}. A general interaction function can then be used to measure the semantic relevance of these embedded KB entries to the question and determine the most likely KB query. 

Most of these approaches use word-level embeddings to encode entities and predicates, and therefore might suffer from the out-of-vocabulary (OOV) problem when they encounter unseen words during test time. Consequently, they often rely on significant data augmentation from sources such as Paralex \cite{fader2013paraphrase}, which contains 18 million question-paraphrase pairs scraped from WikiAnswers, to have sufficient examples for each word they encounter \cite{Bordes2014,qa_acl,babi_dataset}.

As opposed to word-level modeling, character-level modeling can be used to handle the OOV issue. While character-level modeling has not been applied to factoid question answering before, it has been successfully applied to information retrieval, machine translation, sentiment analysis, classification, and named entity recognition \cite{word_hashing,Shen2014,chung2016character,char_level,santos2014learning,dos2014deep,klein2003named,dos2014think,dos2015boosting}. Moreover, \newcite{gf_lstm} demonstrate that gated-feedback LSTMs on top of character-level embeddings can capture long-term dependencies in language modeling.

Lastly, encoder-decoder networks have been applied to many structured machine learning tasks. First introduced in \newcite{sutskever2014sequence}, in an encoder-decoder network, a source sequence is first encoded with a recurrent neural network (RNN) into a fixed-length vector which intuitively captures its \say{meaning}, and then decoded into a desired target sequence. This approach and related memory-based or attention-based approaches have been successfully applied  in diverse domains such as speech recognition, machine translation, image captioning, parsing, executing programs, and conversational dialogues \cite{deep_speech,venugopalan2014translating,align_translate,grammar_language,zaremba2014learning,image_captioning,end2end}.

Unlike previous work, we formulate question answering as a problem of decoding the KB query given the question and KB entries which are encoded in embedding spaces. We therefore integrate the learning of question and KB embeddings in a unified encoder-decoder framework, where the whole system is optimized end-to-end. 

\section{Model}
Since we focus on single-relation question answering in this work, our model decodes every question into a KB query that consists of exactly two elements--the topic entity, and the predicate. More formally, our model is a function $f(q, \{e\}, \{p\})$ that takes as input a question $q$, a set of candidate entities $\{e\}=e_1, ...,e_n$, a set of candidate predicates $\{p\}=p_1,..., p_m$, and produces a likelihood score $p(e_i, p_j|q)$ of generating entity $e_i$ and predicate $p_j$ given question $q$ for all $i\in{1...n}, j\in{1...m}$. 

As illustrated in Figure 1, our model consists of three components:
\begin{enumerate}
  \item A character-level LSTM-based encoder for the question which produces a sequence of embedding vectors, one for each character (Figure 1a). 
  \item A character-level convolutional neural network (CNN)-based encoder for the predicates/entities in a knowledge base which produces a single embedding vector for each predicate or entity (Figure 1c).
  \item An LSTM-based decoder with an attention mechanism and a relevance function for generating the topic entity and predicate to form the KB query (Figure 1b).
\end{enumerate}
The details of each component are described in the following sections. 

\subsection{Encoding the Question}
To encode the question, we take two steps:
\begin{enumerate}
    \item We first extract one-hot encoding vectors for characters in the question, $x_1,...,x_n$, where $x_i$ represents the one-hot encoding vector for the $i^{th}$ character in the question. We keep the space, punctuation and original cases without tokenization. \footnote{We notice that some entity mentions in the question are capitalized and some are not. Also, some questions end with a '?' and some do not.}
    \item We feed $x_1,...,x_n$ from left to right into a two-layer gated-feedback LSTM, and keep the outputs at all time steps as the embeddings for the question, i.e., these are the vectors $s_1,...,s_n$. 
\end{enumerate} 

\subsection{Encoding Entities and Predicates in the KB}
To encode an entity or predicate in the KB, we take two steps:
\begin{enumerate}
    \item We first extract one-hot encoding vectors for characters in its English alias, $x_1,...,x_n$, where $x_i$ represents the one-hot encoding vector for the $i^{th}$ character in the alias. 
    \item We then feed $x_1,...,x_n$ into a temporal CNN with two alternating convolutional and fully-connected layers, followed by one fully-connected layer:
 \begin{align*} 
f(x_1,...,x_n) =  tanh(W_{3} \times max(tanh (W_{2} \times
 \\ 
conv(tanh({W_{1} \times conv(x_1,...,x_n)})))))
\end{align*}
where $f(x_{1...n}) $ is an embedding vector of size $N$, $W_{3}$ has size $R^{N \times h}$, $conv$ represents a temporal convolutional neural network, and $max$ represents a max pooling layer in the temporal direction. 
\end{enumerate}

We use a CNN as opposed to an LSTM to embed KB entries primarily for computational efficiency. Also, we use two different CNNs to encode entities and predicates because they typically have significantly different styles (e.g., \say{Barack Obama} vs. \say{/people/person/place\_of\_birth}).

\subsection{Decoding the KB Query}
To generate the single topic entity and predicate to form the KB query, we use a decoder with two key components:

\begin{enumerate}
    \item An LSTM-based decoder with attention. Its hidden states at each time step $i$, $h_{i}$, have the same dimensionality $N$ as the embeddings of entities/predicates.  The initial hidden state $h_0$ is set to the zero vector: $\vec{0}$.
    
    \item A pairwise semantic relevance function that measures the similarity between the hidden units of the LSTM and the embedding of an entity or predicate candidate. It then returns the mostly likely entity or predicate based on the similarity score. 
\end{enumerate}

In the following two sections, we will first describe the LSTM decoder with attention, followed by the semantic relevance function. 

\subsubsection{LSTM-based Decoder with Attention}
The attention-based LSTM decoder uses a similar architecture as the one described in \newcite{align_translate}. At each time step $i$, we feed in a context vector $c_i$ and an input vector $v_i$ into the LSTM. At time $i=1$ we feed a special input vector $v_{\textless{S}\textgreater}=\vec{0}$ into the LSTM. At time $i=2$, during training, the input vector is the embedding of the true entity, while during testing, it is the embedding of the most likely entity as determined at the previous time step.

We now describe how we produce the context vector $c_i$. Let $h_{i-1}$ be the hidden state of the LSTM at time $i-1$, $s_j$ be the $j^{th}$ question character embedding, $n$ be the number of characters in the question, $r$ be the size of $s_j$, and $m$ be a hyperparameter. Then the context vector $c_i$, which represents the attention-weighted content of the question, is recomputed at each time step $i$ as follows:
\begin{align*}
    c_i =& \sum_{j=1}^{n} \alpha_{ij} s_j,
\end{align*}
\begin{align*}
    \alpha_{ij} =& \frac{\exp\left(e_{ij}\right)}{\sum_{k=1}^{T_x}
    \exp\left(e_{ik}\right)} 
\end{align*}
\begin{align*}
    e_{ij} =& v_a^{\top} \tanh\left( W_a h_{i-1} + U_a s_j \right),
\end{align*}

\noindent where $\{\alpha\}$ is the attention distribution that is applied over each hidden unit $s_j$, $W_a \in R^{m \times N}, U_a \in R^{m \times r},$ and $v_a \in {R}^{1 \times m}$.

\begin{table*}
\begin{center}
\begin{small}
\begin{tabular}{|c|c|c|c|c|c|c|c|c|c|}
\hline
\multicolumn{7}{|l|}{{\sc Results on SimpleQuestions dataset}} &
\multicolumn{3}{l|}{} 
\\ 
\hline
{\sc KB} & \multicolumn{3}{|c|}{\sc Train sources} &
                                           {\sc Autogen.} & {\sc Embed} & {\sc Model} & {\sc Ensemble} & {\sc SQ Accuracy} & {\sc \# Train}\\

 & {\sf WQ} & {\sf SIQ} & {\sf PRP} &  {\sc Questions} & {\sc Type}  & {}  & {}  & {} & {\sc Examples}\\

\hline
FB2M & no & yes & no & no & Char & Ours & 1 model & \bf{70.9} & 76K \\
FB2M & no & yes & no & no & Word
& Ours & 1 model & 53.9 & 76K \\
FB2M & yes & yes & yes & yes & Word & MemNN & 1 model & 62.7 & 26M
\\
\hline
FB5M & no & yes & no & no & Char & Ours & 1 model & {\bf 70.3} & 76K \\
FB5M & no & yes & no & no & Word & Ours & 1 model & 53.1 & 76K \\
FB5M & yes & yes & yes & yes & Word & MemNN & 5 models & 63.9 & 27M \\
FB5M & yes & yes & yes & yes & Word & MemNN & Subgraph & 62.9 & 27M \\
FB5M & yes & yes & yes & yes & Word & MemNN & 1 model & 62.2 & 27M \\
\hline
\end{tabular}

\caption{\label{tab:res} Experimental results on the SimpleQuestions dataset. MemNN results are from \protect\newcite{babi_dataset}. {\sf WQ}, {\sf SIQ} and {\sf PRP} stand for WebQuestions, SimpleQuestions and extracted paraphrases from WikiAnswers, respectively.}
\end{small}
\end{center}
\vspace*{-1ex}
\end{table*}

\subsubsection{Semantic Relevance Function}
Unlike machine translation and language modeling where the vocabulary is relatively small, there are millions of entries in the KB. If we try to directly predict the KB entries, the decoder will need an output layer with millions of nodes, which is computationally prohibitive. Therefore, we resort to a relevance function that measures the semantic similarity between the decoder's hidden state and the embeddings of KB entries. Our semantic relevance function takes two vectors $x_1$, $x_2$ and returns a distance measure of how similar they are to each other. In current experiments we use a simple cosine-similarity metric: $cos(x_1, x_2)$. 

Using this similarity metric, the likelihoods of generating entity $e_j$ and predicate $p_k$ are:

\begin{align*}
\hspace*{0.0cm}
P(e_j) = \frac{exp(\lambda cos(h_1,e_{j}))}{\sum_{i=1}^{n} exp(\lambda cos(h_1,e_i))}
\\ 
P(p_k) = \frac{exp(\lambda cos(h_2,p_{k}))}{\sum_{i=1}^{m} exp(\lambda cos(h_2,p_{i}))}
\end{align*}
where $\lambda$ is a constant, $h_1, h_2$ are the hidden states of the LSTM at times $t=1$ and $t=2$, $e_1,...,e_n$ are the entity embeddings, and $p_1,...,p_m$ are the predicate embeddings. A similar likelihood function was used to train the semantic similarity modules proposed in \newcite{qa_acl} and \newcite{Yih2015SemanticPV}. 

During inference, $e_1,...,e_n$ and $p_1,...,p_m$ are the embeddings of candidate entities and predicates. During training $e_1,...,e_n$, $p_1,...,p_m$ are the embeddings of the true entity and 50 randomly-sampled entities, and the true predicate and 50 randomly-sampled predicates, respectively.

\subsection{Inference}
For each question $q$, we generate a candidate set of entities and predicates, $\{e\}$ and $\{p\}$, and feed it through the model $f(q, \{e\}, \{p\})$. We then decode the most likely (entity, predicate) pair:
\begin{align*}
(e^*, p^*) = argmax_{e_i, p_j} (P(e_i)*P(p_j))
\end{align*} which becomes our semantic parse.

We use a similar procedure as the one described in \newcite{babi_dataset} to generate candidate entities $\{e\}$ and predicates $\{p\}$. Namely, we take all entities whose English alias is a substring of the question, and remove all entities whose alias is a substring of another entity. For each English alias, we sort each entity with this alias by the number of facts that it has in the KB, and append the top 10 entities from this list to our set of candidate entities. All predicates ${p_j}$ for each entity in our candidate entity set become the set of candidate predicates.

\subsection{Learning}
Our goal in learning is to maximize the joint likelihood $P(e_c) \cdot P(p_c)$ of predicting the correct entity $e_c$ and predicate $p_c$ pair from a set of randomly sampled entities and predicates. We use back-propagation to learn all of the weights in our model. 

All the parameters of our model are learned jointly without pre-training. These parameters include the weights of the character-level embeddings, CNNs, and LSTMs. Weights are randomly initialized before training. For the $i^{th}$ layer in our network, each weight is sampled from a uniform distribution between $-\frac{1}{|l^i|}$ and $\frac{1}{|l^i|}$, where $|l^i|$ is the number of weights in layer $i$.

\section{Dataset and Experimental Settings} \label{word_count}
We evaluate the proposed model on the SimpleQuestions dataset \cite{babi_dataset}. The dataset consists of 108,442 single-relation questions and their corresponding (\textit{topic entity}, \textit{predicate}, \textit{answer entity}) triples from Freebase. It is split into 75,910 train, 10,845 validation, and 21,687 test questions. Only 10,843 of the 45,335 unique words in entity aliases and 886 out of 1,034 unique predicates in the test set were present in the train set. For the proposed dataset, there are two evaluation settings, called FB2M and FB5M, respectively. The former uses a KB for candidate generation which is a subset of Freebase and contains 2M entities, while the latter uses subset of Freebase with 5M entities.

In our experiments, the Memory Neural Networks (MemNNs) proposed in \newcite{babi_dataset} serve as the baselines. For training, in addition to the 76K questions in the training set, the MemNNs use 3K training questions from WebQuestions \cite{Berant:EMNLP13}, 15M paraphrases from WikiAnswers \cite{fader2013paraphrase}, and 11M and 12M automatically generated questions from the KB for the FB2M and FB5M settings, respectively. In contrast, our models are trained only on the 76K questions in the training set.

For our model, both layers of the LSTM-based question encoder have size 200. The hidden layers of the LSTM-based decoder have size 100, and the CNNs for entity and predicate embeddings have a hidden layer of size 200 and an output layer of size 100. The CNNs for entity and predicate embeddings use a receptive field of size 4, $\lambda=5$, and $m=100$. We train the models using RMSProp with a learning rate of $1e^{-4}$.

In order to make the input character sequence long enough to fill up the receptive fields of multiple CNN layers, we pad each predicate or entity using three padding symbols $P$, a special start symbol, and a special end symbol. For instance, $Obama$ would become $S_{start}PPP ObamaPPPS_{end}$. For consistency, we apply the same padding to the questions.

\noindent
    

\section{Results}
\begin{table*}
\begin{small}
\begin{center}
\resizebox{1\linewidth}{!}{
\begin{tabular}{llllll}
\multicolumn{1}{c}{\bf \# of LSTM Layers}
&\multicolumn{1}{c}{\bf \# of CNN Layers}
&\multicolumn{1}{c}{\bf Joint Accuracy}
&\multicolumn{1}{c}{\bf Predicate Accuracy}
&\multicolumn{1}{c}{\bf Entity Accuracy}
\\ \hline
2 & 2 & \bf{78.3} & \bf{80.0} & 96.6 \\
2 & 1 & 77.7 & 79.4 & \bf{96.8} \\
1 & 2 & 71.5 & 73.9 & 95.0 \\
1 & 1 & 72.2 & 74.7 & 94.9
\end{tabular}
}

\caption{\label{tab:vis} \small Results for a random sampling experiment where we varied the number of layers used for convolutions and the question-encoding LSTM. We terminated training models after 14 epochs and 3 days on a GPU.}

\end{center}
\end{small}
\end{table*}

\begin{table*}
\begin{small}
\begin{center}
\begin{tabular}{lllll}
\multicolumn{1}{c}{\bf Embedding Type}
&\multicolumn{1}{c}{\bf Joint Accuracy}
&\multicolumn{1}{c}{\bf Predicate Accuracy}
&\multicolumn{1}{c}{\bf Entity Accuracy}
\\ \hline
Character & \bf{78.3} & \bf{80.0} & \bf{96.6} \\
Word & 37.6 & 78.8 & 45.5
\end{tabular}
\caption{\label{tab:vis} \small Results for a random sampling experiment where we varied the embedding type (word vs. character-level). We used 2 layered-LSTMs and CNNs for all our experiments. Our models were trained for 14 epochs and 3 days on a GPU.}
\end{center}
\end{small}
\end{table*}

\subsection{End-to-end Results on SimpleQuestions}
Following \newcite{babi_dataset}, we report results on the SimpleQuestions dataset in terms of SQ accuracy, for both FB2M and FB5M settings in Table 1. SQ accuracy is defined as the percentage of questions for which the model generates a correct KB query (i.e., both the topic entity and predicate are correct). Our single character-level model achieves SQ accuracies of 70.9\% and 70.3\% on the FB2M and FB5M settings, outperforming the previous state-of-art results by 8.2\% and 6.4\%, respectively. Compared to the character-level model, which only has 1.2M parameters, our word-level model has 19.9M parameters, and only achieves a best SQ accuracy of 53.9\%. In addition, in contrast to previous work, the OOV issue is much more severe for our word-level model, since we use no data augmentation to cover entities unseen in the train set.

\subsection{Ablation and Embedding Experiments}
We carry out ablation studies in Sections 5.2.1 and 5.2.2 through a set of \textit{random-sampling} experiments. In these experiments, for each question, we randomly sample 200 entities and predicates from the test set as noise samples. We then mix the gold entity and predicate into these negative samples, and evaluate the accuracy of our model in predicting the gold predicate or entity from this mixed set. 

\subsubsection{Character-Level vs. Word-Level Models} 
We first explore using word-level models as an alternative to character-level models to construct embeddings for questions, entities and predicates.

Both word-level and character-level models perform comparably well when predicting the predicate, reaching an accuracy of around 80\% (Table 3). However, the word-level model has considerable difficulty generalizing to unseen entities, and is only able to predict 45\% of the entities accurately from the mixed set. These results clearly demonstrate that the OOV issue is much more severe for entities than predicates, and the difficulty word-level models have when generalizing to new entities.

In contrast, character-level models have no such issues, and achieve a 96.6\% accuracy in predicting the correct entity on the mixed set. This demonstrates that character-level models encode the semantic representation of entities and can match entity aliases in a KB with their mentions in natural language questions.

\subsubsection{Depth Ablation Study}
We also study the impact of the depth of neural networks in our model. The results are presented in Table 2. In the ablation experiments we compare the performance of a single-layer LSTM to a two-layer LSTM to encode the question, and a single-layer vs. two-layer CNN to encode the KB entries. We find that a two-layer LSTM boosts joint accuracy by over 6\%. The majority of accuracy gains are a result of improved predicate predictions, possibly because entity accuracy is already saturated in this experimental setup.
\subsection{Attention Mechanisms}
In order to further understand how the model performs question answering, we visualize the attention distribution over question characters in the decoding process. In each sub-figure of Figure 2, the x-axis is the character sequence of the question, and the y-axis is the attention weight distribution $\{\alpha_i\}$. The blue curve is the attention distribution when generating the entity, and green curve is the attention distribution when generating the predicate.

Interestingly, as the examples show, the attention distribution typically peaks at empty spaces. This indicates that the character-level model learns that a space defines an ending point of a complete linguistic unit. That is, the hidden state of the LSTM encoder at a space likely summarizes content about the character sequence before that space, and therefore contains important semantic information that the decoder needs to attend to.

Also, we observe that entity attention distributions are usually less sharp and span longer portions of words, such as \say{john} or \say{rutters}, than predicate attention distributions (e.g., Figure 2a). For entities, semantic information may accumulate gradually when seeing more and more characters, while for predicates, semantic information will become clear only after seeing the complete word. For example, it may only be clear that characters such as \say{song by} refer to a predicate after a space, as opposed to the name of a song such as \say{song bye bye love} (Figures 2a, 2b). In contrast, a sequence of characters starts to become a likely entity after seeing an incomplete name such as \say{joh} or \say{rutt}. 

In addition, a character-level model can identify entities whose English aliases were never seen in training, such as \say{phrenology} (Figure 2d). The model apparently learns that words ending with the suffix \say{nology} are likely entity mentions, which is interesting because it reads in the input one character at a time. 

Furthermore, as observed in Figure 2d, the attention model is capable of attending disjoint regions of the question and capture the mention of a predicate that is interrupted by entity mentions. We also note that predicate attention often peaks at the padding symbols after the last character of the question, possibly because sentence endings carry extra information that further help disambiguate predicate mentions. In certain scenarios, the network may only have sufficient information to build a semantic representation of the predicate after being ensured that it reached the end of a sentence. 

Finally, certain words in the question help identify both the entity and the predicate. For example, consider the word \say{university} in the question \say{What type of educational institution is eastern new mexico university} (Figure 2c). Although it is a part of the entity mention, it also helps disambiguate the predicate. However, previous semantic parsing-based QA approaches \cite{Yih2015SemanticPV,qa_acl} assume that there is a clear separation between the predicate and entity mentions in the question. In contrast, the proposed model does not need to make this hard categorization, and attends the word \say{university} when predicting both the entity and predicate.

\section{Error Analysis}
We randomly sampled 50 questions where the best-performing model generated the wrong KB query and categorized the errors. For 46 out of the 50 examples, the model predicted a predicate with a very similar alias to the true predicate, i.e. \say{/music/release/track} vs. \say{/music/release/track\_list}. For 21 out of the 50 examples, the model predicted the wrong entity, e.g., \say{Album} vs. \say{Still Here} for the question \say{What type of album is still here?}. Finally, for 18 of the 50 examples, the model predicted the wrong entity and predicate, i.e. (\say{Play}, \say{/freebase/equivalent\_topic/equivalent\_type}) for the question \say{which instrument does amapola cabase play?} Training on more data, augmenting the negative sample set with words from the question that are not an entity mention, and having more examples that disambiguate between similar predicates may ameliorate many of these errors.

\section{Conclusion}

In this paper, we proposed a new character-level, attention-based encoder-decoder model for question answering. In our approach, embeddings of questions, entities, and predicates are all jointly learned to directly optimize the likelihood of generating the correct KB query. Our approach improved the state-of-the-art accuracy on the SimpleQuestions benchmark significantly, using much less data than previous work. Furthermore, thanks to character-level modeling, we have a compact model that is robust to unseen entities. Visualizations of the attention distribution reveal that our model, although built on character-level inputs, can learn higher-level semantic concepts required to answer a natural language question with a structured KB. In the future we would like to extend our system to handle multi-relation questions.


\begin{figure}
\small
\begin{center}
\begin{tabular}{l}


\raisebox{11\height}{a)} \includegraphics[scale=0.325]{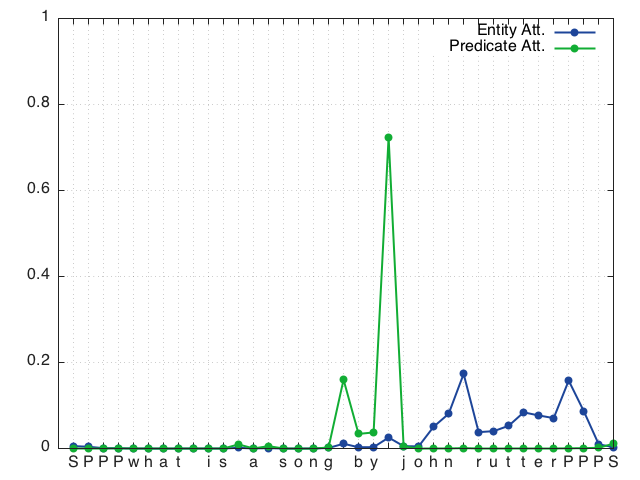} \\
\raisebox{11\height}{b)} \includegraphics[scale=0.325]{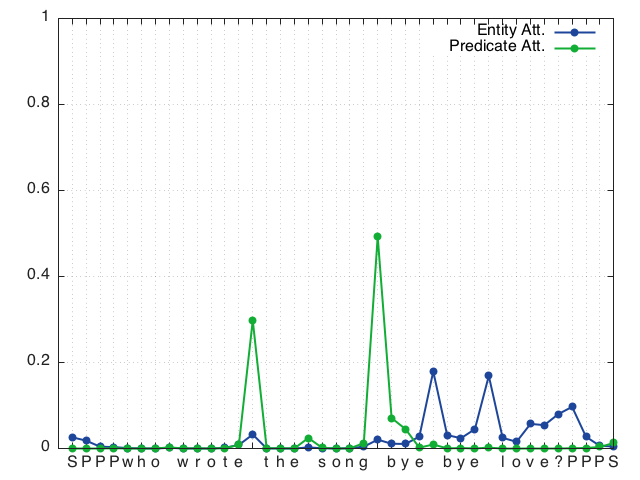} \\
\raisebox{11\height}{c)} \includegraphics[scale=0.325]{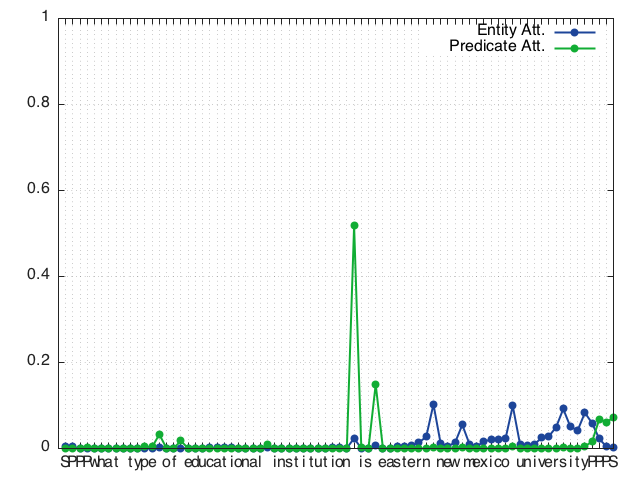} \\
\raisebox{11\height}{d)} \includegraphics[scale=0.325]{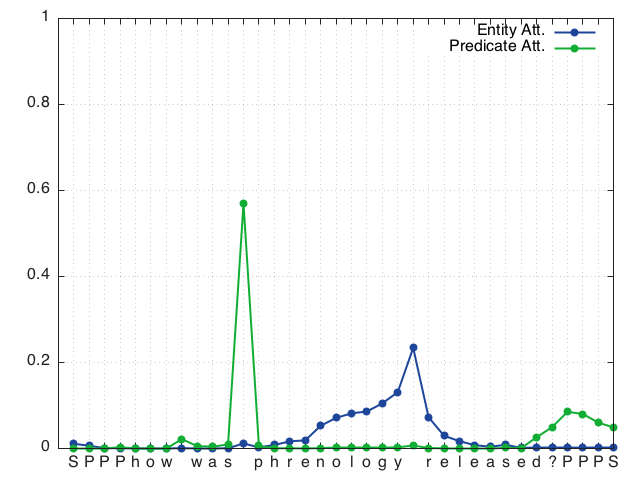}
\end{tabular}

\caption{\label{tab:vis} Attention distribution over outputs of a left-to-right LSTM on question characters.}

\end{center}
\end{figure}

\clearpage
\bibliography{emnlp2016}
\bibliographystyle{emnlp2016}

\end{document}